\documentclass{article}

\usepackage{microtype}
\usepackage{graphicx}
\usepackage{subcaption}
\usepackage{booktabs} 

\usepackage{hyperref}

\usepackage[accepted]{icml2023}

\usepackage{amsmath}
\usepackage{amssymb}
\usepackage{mathtools}
\usepackage{amsthm}
\usepackage{bm}

\usepackage[capitalize,noabbrev]{cleveref}

\theoremstyle{plain}

\theoremstyle{definition}

\theoremstyle{remark}

\usepackage[textsize=tiny]{todonotes}

\icmltitlerunning{Conservative objective models are a special kind of contrastive divergence-based energy model}

\begin{document}

\onecolumn
\icmltitle{Conservative objective models are a special kind of contrastive divergence-based energy model}



\icmlsetsymbol{equal}{*}

\begin{icmlauthorlist}
\icmlauthor{Christopher Beckham}{mila,polymtl}
\icmlauthor{Christopher Pal}{mila,polymtl,sn,cifar}
\end{icmlauthorlist}

\icmlaffiliation{mila}{Mila - Quebec Artificial Intelligence Institute}
\icmlaffiliation{polymtl}{Polytechnique Montreal}
\icmlaffiliation{sn}{ServiceNow Research}
\icmlaffiliation{cifar}{CIFAR AI Chair}
\icmlcorrespondingauthor{Christopher Beckham}{first.last@mila.quebec}

\icmlkeywords{Machine Learning, model-based optimisation, model-based optimization, conservative objective models, energy-based model, contrastive divergence, score matching, generative models, generative models for design, gflownets}

\vskip 0.3in



\printAffiliationsAndNotice{}  

\newcommand{\pt}{p_{\theta}}
\newcommand{\pomega}{p_{\omega}}

\newcommand{\pphi}{p_{\phi}}
\newcommand{\ptp}{p_{\theta, \pi}}
\newcommand{\ptphi}{p_{\theta, \phi}}
\newcommand{\ptk}{p^{k}_{\theta}}
\newcommand{\ft}{f_{\theta}}
\newcommand{\fomega}{f_{\omega}}
\newcommand{\Et}{E_{\theta}}
\newcommand{\st}{s_{\theta}}
\newcommand{\Zt}{Z_{\theta}}
\newcommand{\Lt}{\mathcal{L}_{\theta}}
\newcommand{\Lsup}{\mathcal{L}^{\text{sup}}_{\theta}}
\newcommand{\Lunsup}{\mathcal{L}^{\text{unsup}}_{\theta}}
\newcommand{\Ltp}{\mathcal{L}_{\theta, \pi}}
\newcommand{\prior}{p_{\pi}}
\newcommand{\argmax}{\text{argmax}}
\newcommand{\argmin}{\text{argmin}}
\newcommand{\xx}{\bm{x}}
\newcommand{\zz}{\bm{z}}

\begin{abstract}
In this work we theoretically show that conservative objective models (COMs) for offline model-based optimisation (MBO) are a special kind of contrastive divergence-based energy model, one where the energy function represents both the unconditional probability of the input and the conditional probability of the reward variable. While the initial formulation only samples modes from its learned distribution, we propose a simple fix that replaces its gradient ascent sampler with a Langevin MCMC sampler. This gives rise to a special probabilistic model where the probability of sampling an input is proportional to its predicted reward. Lastly, we show that better samples can be obtained if the model is decoupled so that the unconditional and conditional probabilities are modelled separately.
\end{abstract}

\section{Introduction}

Model-based optimisation (MBO) is concerned with the use of generative models for design problems, where the input $\xx$ specifies the design and the desirability of any design (i.e. the reward) is a black box function $y = f(\xx)$ called the \emph{ground truth oracle} which is prohibitively expensive to evaluate. For instance, if we are dealing with designing drugs to target disease then the oracle is a real world process that involves synthesising and testing the drug in a wet lab, which is expensive. Because each evaluation of the `real world' $f(\xx)$ is expensive, we would like to use machine learning to construct a reliable proxy of the oracle $\ft(\xx)$ and exploit that instead. (This is one discernible difference to more traditional derivative-free black box optimisation, which assumes that $f$ can be queried at will.) In addition, we are also interested in \emph{extrapolation}: we would like to find designs $\xx$ that have as high of a reward as possible, possibly even higher than what has been observed so far. Generally speaking, we would like to generate a candidate set $\mathcal{S} = \{ \xx_j \}_{j=1}^{M}$ such that:
\begin{align} \label{eq:mbo}
\mathcal{S} = \argmax_{\xx_1, \dots, \xx_M} \ \sum_{j=1}^{M} f(\xx_j). 
\end{align}
Since we do not have access to $f$ during training, we must resort to training an approximation of it, which we denote $\ft(\xx)$. This is usually called an \emph{approximate} or \emph{surrogate} model or `oracle'. We note that because $\ft(\xx)$ is approximate and is a discriminative model, it is vulnerable to over-scoring inputs or even assigning rewards greater than zero to implausible inputs,\footnote{By `implausible' inputs, we simply mean any input for which $p(\xx) = 0$.} and these are commonly referred to as \emph{adversarial examples}. In the context of our aforementioned drug example, an implausible input would be a drug whose chemical configuration (that is, its configuration of atoms) is physically impossible. How these problems are addressed depends on whether one approaches MBO from a discriminative modelling point of view \cite{fu2021offline,trabucco2021conservative} or a generative modelling one \cite{brookes2019conditioning,fannjiang2020autofocused,kumar2020model,beckham2022towards}. In this work we will exclusively discuss \emph{conservative objective models} \cite{trabucco2021conservative}, whose work comes from the discriminative perspective. However, we will show that this model essentially falls under a particular class of generative model called an \emph{energy-based model}, and in this work we perform a theoretical and empirical analysis of that model from this perspective.

Lastly, for the sake of clarification, we note that MBO methods can be categorised into whether they are online or offline. In the online case, we assume that the ground truth oracle can be queried during training to obtain additional labels, which essentially becomes active learning. In the offline case, we only assume a dataset $\mathcal{D} = \{(\xx_i, y_i)\}_{i=1}^{n}$ and must make do with this data to train the best possible proxy model $\ft(\xx)$.\footnote{The difference between offline and online MBO pertains to just the training of the generative model. Even with `offline' MBO, in a real world setting that model still has to be validated against the ground truth oracle by generating novel inputs and scoring them.} For the remainder of this paper we will only consider offline MBO, and simply refer to it as MBO unless otherwise stated.

We lay out our contributions in this work as follows:
\begin{itemize}
    \item We theoretically show that conservative objective models (COMs) are extremely similar to an energy-based model (EBM) that is trained via contrastive divergence, albeit with a modified MCMC sampler that can only sample from the \emph{modes} of its distribution (Section \ref{sec:connection}). This special form of EBM is parameterised such that the \emph{negative energy} of an input $-\Et(\xx)$ is equivalent to the predicted reward $\ft(\xx)$ of that same input. In other words, the energy is trained to be both predictive of the \emph{likelihood} of an example, as well as its \emph{reward}, with a training hyperparameter $\alpha$ introduced to balance the trade-off between how much model capacity should be allocated between the two. These two components can be seen as inducing a \emph{joint density} $\pt(\xx,y; \alpha) \propto \pt(y|\xx)\pt(\xx)^{\alpha}$ over the data.\footnote{Code for this paper will be made available here: \url{https://github.com/christopher-beckham/coms-are-energy-models}}
    \item COMs uses gradient ascent for its MCMC sampler, which can only sample modes from its distribution. If it is modified to properly sample from the distribution, then the model becomes a special instance of a contrastive divergence EBM, and we call these \emph{\textbf{Stochastic COMs}} (Section \ref{sec:coms_stochastic}). \emph{\textbf{Stochastic COMs}} have the special property that the probability of sampling an input is proportional to its predicted reward, i.e. $\pt(\xx) \propto \ft(\xx)$. We illustrate the effect of $\alpha$ on a toy spiral dataset in Section \ref{sec:toy} as well as visualise generated samples between the both COMs variants.
    \item We show that COMs fail to generate desirable samples on a simple toy spiral dataset because the same network is being used to parameterise both the likelihood of an example and its score, and this subsequently degrades sample quality. To alleviate this, we propose \emph{\textbf{decoupled COMs}}, where a separate classifier is trained and its gradients are used at sampling time (Section \ref{sec:decoupled}). \emph{\textbf{Decoupled COMs}} can be thought of as a contrastive divergence-based EBM which leverages an external classifier (regression model) as a form of conditional guidance.
\end{itemize}


\subsection{Energy-based generative models}

In EBMs we wish to learn a probability distribution without any specific modelling assumptions. This is done by defining the unnormalised probability of an input as the \emph{negative} of an energy $\Et$ that is parameterised with a neural network:
\begin{align} \label{eq:eq1}
\pt(\xx) = \frac{\exp(-\Et(\xx))}{\Zt}, 
\end{align}
where $\Zt$ is the (usually intractable) normalising constant, which can be seen as a function of the energy model parameters $\theta$. Ignoring the intractability issue for a brief moment, the log likelihood for one example $\xx$ can be expressed as:
\begin{align} \label{eq:log_px}
\log \pt(\xx) = -\Et(\xx) - \log \underbrace{\int_{\xx} \exp(-\Et(\xx)) d\xx}_{\Zt},
\end{align}
where we have re-written $\Zt$ as an integral. While this seems virtually impossible to handle, an interesting identity from \citet{ebms_tutorial} says the score of $\Zt$ is equivalent to:
\begin{align} \label{eq:z_trick}
\nabla_{\theta} \log \Zt = \mathbb{E}_{\xx \sim \pt(\xx)} [ -\nabla_{\xx} \Et(\xx) ].
\end{align}
In other words, the integral can be approximated via Monte Carlo by simply computing the score over each example inside the expected value. This means that we can define a loss $\Lt(\xx)$ such that, when we take the gradient of it, it becomes equivalent to $\nabla_{\theta} \log \pt(\xx)$:
\begin{align} \label{eq:integral_trick}
\Lt(\xx)& = -\Et(\xx) + \mathbb{E}_{\xx \sim \pt(\xx)} \Et(\xx) \nonumber \\
\implies \nabla_{\theta} \Lt(\xx) = \nabla_{\theta} \log \pt(\xx) & = \nabla_{\theta}[-\Et(\xx)] + \underbrace{\mathbb{E}_{\xx \sim \pt(\xx)} [ -\nabla_{\xx} \Et(\xx) ]}_{\text{Eqn. \ref{eq:z_trick}}}.
\end{align}
It is expensive to approximate $\Zt$ term because it requires us to draw samples from the generative model $\pt(\xx)$ which is a costly process. For example, we would have to run Langevin MCMC \cite{neal2011mcmc,welling2011bayesian,ebms_tutorial} by drawing an initial $\xx_0$ from some simple prior distribution and running the Markov chain for a sufficiently long number of time steps $T$ such that $\xx_T \approx \pt(\xx)$:
\begin{align} \label{eq:langevin_mcmc}
\xx_{t+1} & := \xx_{t} + \frac{\epsilon_t^2}{2} \nabla_{\xx_t} \log \pt(\xx_t) + \epsilon \zz_{t} \nonumber \\
\Aboxed{& = \xx_{t} + \frac{\epsilon_t^2}{2} \nabla_{\xx_t}[-\Et(\xx_t)] + \epsilon_t \zz_{t},}
\end{align}
where $\zz_t \sim \mathcal{N}(0, \bm{I})$, and  $\epsilon_{T} \rightarrow 0$.

For reasons that will become clear shortly, we prefer to define $\pt(\xx)$ more explicitly such that it is obvious that sampling involves an $\xx_0$ that is drawn from a simple prior distribution (e.g. a Gaussian or uniform distribution), which we will call $\prior(\xx)$. That is, in order to sample from our generative model we first sample $\xx_0 \sim \prior(\xx_0)$ and then run Langevin MCMC on $\xx_0$, which we can write simply as a sample from the conditional distribution $\xx \sim \pt(\xx|\xx_0)$. Both of these distributions define a joint distribution $\ptp(\xx, \xx_0) = \pt(\xx|\xx_0)\prior(\xx_0)$, and therefore the marginal over $\xx$ itself can simply be written as:
\begin{align}
\ptp(\xx) = \int_{\xx_0} \pt(\xx|\xx_0)\prior(\xx_0) d \xx_0.
\end{align}
Therefore, we can write a more explicit form of Equation \ref{eq:log_px} that uses $\ptp$ instead:
\begin{align} \label{eq:log_px_ebm}
    \Ltp(\xx) = \log \ptp(\xx) & = -\Et(\xx) + \mathbb{E}_{\xx' \sim \ptp(\xx) } \Et(\xx') \nonumber \\
    & = -\Et(\xx) + \mathbb{E}_{\xx' \sim \pt(\xx|\xx_0), \xx_0 \sim \prior(\xx_0)} \Et(\xx').
\end{align}
We now explain the reason for this reformulation: a widely known algorithm used to train these models is called \emph{contrastive divergence} \citep{hinton2002training}, a modification of the Langevin MCMC procedure. Contrastive divergence proposes two modifications to make it more computationally viable: run MCMC for $k$ iterations instead (where $k$ is extremely small, such as a few steps), and let $\prior$ be the \emph{actual data distribution}, so the chain is initialised from a real data point. (To keep notation simple, whenever $\pt(\xx)$ is used, we really mean $\ptp(\xx)$ where $p_{\pi}(\xx) = p(\xx)$, the real data distribution.) While running the sampling chain for $k$ iterations introduces some bias into the gradient, it appears to work well in practice \cite{bengio2009justifying}. Concretely, if we use contrastive divergence for $k$ iterations then we will write our objective as the following:
\begin{align} \label{eq:cd_k}
    \Lt^{CD-k} (\xx) & := -\Et(\xx) + \mathbb{E}_{\xx' \sim \ptk(\xx'|\xx_0)p(\xx_0)} \Et(\xx') \\
    & \approx \Lt(\xx). \nonumber
\end{align}

We will denote this style of energy-based model as a \emph{\textbf{contrastive divergence-based EBM}}, or simply \emph{\textbf{CD-EBM}}.



\section{Conservative objective models} \label{sec:coms}

Before continuing, we make an important distinction between the approximate oracle $\ft(\xx)$ itself and its \emph{statistical} interpretation, $\pt(y|\xx)$. The approximate oracle $\ft(\xx)$ is a regression model trained to predict $y$ from $\xx$ but the precise loss function used imbues a specific probabilistic interpretation relating to that model. For instance, if the \emph{mean squared error loss} is used during training, then $\pt(y|\xx)$ has the interpretation of being a Gaussian distribution whose $\ft(\xx)$ parameterises the mean and $\sigma^2 = 1$. While the choice of probabilistic model is up to the user, we will assume a Gaussian model here as it is the most commonly used for regression tasks and is the probabilistic model used in the paper. Given some training pair $(\xx,y) \in \mathcal{D}$ we can write out its conditional likelihood of $y$ given $\xx$ as follows:
\begin{align} \label{eq:pyx}
\log \pt(y|\xx) = \log \mathcal{N}(y; \ft(\xx), \sigma) = -\frac{1}{\sigma\sqrt{2\pi}}(y - \ft(\xx))^2,
\end{align}
and since we assumed $\sigma^2 = 1$ we get $-(y - \ft(\xx))^2$ times a constant term. Since the mean squared error loss is typically minimised, the negative sign disappears. 

Conservative objective models (COMs) are a recently proposed method \citep{trabucco2021conservative} for MBO. Conceptually, the method can be thought of as simply training an approximate oracle $\ft(\xx)$ but with the model subjected to an extra regularisation term that penalises predictions for samples that have been generated with $\ft$, which are assumed to be adversarial examples. In order to mitigate the issue of adversarial examples and over-scoring, the authors propose a regularisation term that penalises the magnitude of samples that have been generated in the vicinity of $\xx$:
\begin{align} \label{eq:coms_loss}
    \Lsup(\xx,y; \alpha) := \log \pt(y|\xx) + \underbrace{\alpha\big[ - \mathbb{E}_{\xx' \approx \pt(\xx'|\xx_0), \xx_0 \sim p(\xx)} \ft(\xx') + \ft(\xx) \big]}_{\text{COMs regulariser}}.
\end{align}
The following sampler is used for $\pt(\xx|\xx_0)$:
\begin{align} \label{eq:grad_asc}
\xx_{t+1} & := \xx_{t} + \epsilon \nabla_{\xx_t}[ -\Et(\xx_t) ] \nonumber \\
\Aboxed{ & = \xx_{t} + \epsilon \nabla_{\xx_t} \ft(\xx_t),}
\end{align} 
where $\epsilon$ is constant for each time step. What is interesting is that this procedure does not inject any noise; because of this, samples will instead converge to a \emph{maximum a posteriori} solution, i.e. one of the modes of the distribution $\pt(\xx)$ \cite{welling2011bayesian} (hence the use of the approximate symbol $\approx$ in the expectation of Equation \ref{eq:grad_asc}). This can be problematic if there is very little inter-sample diversity amongst generated samples, as they will be less robust as a whole to the ground truth oracle.





\subsection{Relationship to EBMs} \label{sec:connection}

Furthermore, we note that the regularisation term inside $\alpha$ in Equation \ref{eq:coms_loss} is actually \emph{equivalent} to Equation \ref{eq:log_px_ebm} if we define $\ft(\xx) = -\Et(\xx)$, and this in turn is equivalent to $\log \pt(\xx)$. This, combined with the classification loss $\log \pt(y|\xx)$ defines a \emph{joint distribution} $\pt(\xx,y)$, which is precisely the loss proposed in the original paper \cite{trabucco2021conservative}. Let us propose a special joint density $p(\xx, y; \alpha)$ where $\alpha$ controls the trade-off between the two likelihood terms:
\begin{align}
\pt(\xx, y; \alpha) & \propto \pt(y|\xx) \pt(\xx)^{\alpha} \nonumber \\
\implies \log \pt(\xx, y; \alpha) & \propto \log \pt(y|\xx) + \alpha \log \pt(\xx) \label{eq:joint_coms} \\
& = \underbrace{-\frac{1}{2}(y - \ft(\xx))^2}_{\log \pt(y|\xx)} + \alpha\big[ - \mathbb{E}_{\xx' \sim \pt(\xx'|\xx_0), \xx_0 \sim p(\xx)} \ft(\xx') + \ft(\xx) \big] \nonumber \\
& = -\frac{1}{2}(y - \underbrace{\ft(\xx)}_{-\Et(\xx)})^2 + \alpha\underbrace{\big[ \mathbb{E}_{\xx' \sim \pt(\xx)} \Et(\xx') -\Et(\xx) \big]}_{\text{$\log \pt(\xx)$, Eqn. \ref{eq:integral_trick}}}
\end{align}
Setting aside for now the fact that Equation \ref{eq:grad_asc} is not properly sampling from $\pt(\xx)$, what we are observing is a special type of \emph{CD-EBM} where the \emph{negative energy} $-\Et(\xx)$ is equivalent to the predicted reward $\ft(\xx)$. In other words, $\pt(\xx)$ is \emph{proportional} to $\ft(\xx)$. The coefficient $\alpha$ dictates the balance between the classification loss and the marginal likelihood over $\xx$. For instance, if $\alpha = 0$ then $\log p(y|\xx)$ remains and no density estimation is being done over $\xx$. Conversely, if $\alpha$ was extremely large then it would not be a good predictor of the actual reward of $\xx$, but good at modelling the distribution of $\xx$'s since the model is heavily skewed to favour that task. Intuitively then, it would seem that one would want to choose an $\alpha$ such that both tasks are performed well, but this may be difficult to achieve. We will return to this issue in Section \ref{sec:decoupled}.


\subsection{\emph{Stochastic} COMs using Langevin MCMC} \label{sec:coms_stochastic}

Previously, we mentioned that COMs' sampling procedure is not actually drawing samples from the generative model $\pt(\xx)$; instead, it is simply finding a \emph{maximum a posteriori} solution (i.e. a mode of the distribution). To fix this, we simply need to replace the gradient ascent sampler with Langevin MCMC algorithm in Equation \ref{eq:langevin_mcmc}:
\begin{align} \label{eq:langevin_mcmc_coms_uncond}
    \xx_{t+1} & := \xx_{t} + \frac{\epsilon_t^2}{2} \nabla_{\xx_t} \log \pt(\xx_t) + \epsilon_t \zz_{t} \nonumber \\
    \Aboxed{& = \xx_{t} + \frac{\epsilon_t^2}{2} \nabla_{\xx_t}\underbrace{\ft(\xx_t)}_{-\Et(\xx)} + \epsilon_t \zz_{t},}
\end{align}


\subsection{\emph{Decoupled} COMs} \label{sec:decoupled}

In Section \ref{sec:connection} we showed that COM's training objective induces a special joint density $\pt(\xx, y; \alpha)$ where $\alpha$ controls the trade-off between modelling $\pt(\xx)$ and also the classifier $\pt(y|\xx)$. If $\alpha$ were to be carefully tuned then we would hope that samples drawn from the model $\xx \sim \pt(\xx)$ would not only be plausible (i.e. lie on the data distribution) but also comprise high reward on average. One concern is that since the same model $\Et$ is parameterising both distributions, achieving this might be cumbersome. Here we propose an alternative, one that decouples the training of both. Let us denote $\pt(\xx)$ any learned energy-based model\footnote{We can even use a COM for which $\alpha$ is large enough such that most of the model is spent on modelling the data distribution. In fact, we found that the training dynamics of this was more stable than the training of a CD-EBM.} on $\xx$, and also introduce an independently trained oracle $\fomega(\xx)$ which is a standard regression model trained to predict $y$ from $\xx$. We propose the following tilted density \citep{asmussen2007stochastic, o2020making}:
\begin{align}
p_{\theta,\omega}(\xx; w) & = \pt(\xx)\exp(w\fomega(\xx) - \kappa(1/w)) \\
\implies \log p_{\theta,\omega}(\xx; w) & = \log \pt(\xx) + w\fomega(\xx) - \text{const.},
\end{align}
where $w$ is a hyperparameter weighting our preference for $\xx$'s with high reward (with respect to $\fomega$) and $\kappa$ is a normalising constant and does not depend on $\xx$. To sample, we simply use the following Langevin MCMC sampler:
\begin{align} \label{eq:langevin_mcmc_decoupled}
    \xx_{t+1} & := \xx_{t} + \frac{\epsilon^2}{2} \nabla_{\xx_t} \Big[ w \fomega(\xx) + \log \pt(\xx_t) \Big] + \epsilon \zz_{t} \nonumber \\
    \Aboxed{& = \xx_{t} + \frac{\epsilon^2}{2} \Big[ w \nabla_{\xx_t} \fomega(\xx)  + \nabla_{\xx_t} \ft(\xx_t) \Big] + \epsilon \zz_{t}.}
\end{align}


\section{Experiments and Discussion} \label{sec:toy}

\begin{figure}
    \centering
    \includegraphics[width=0.7\textwidth]{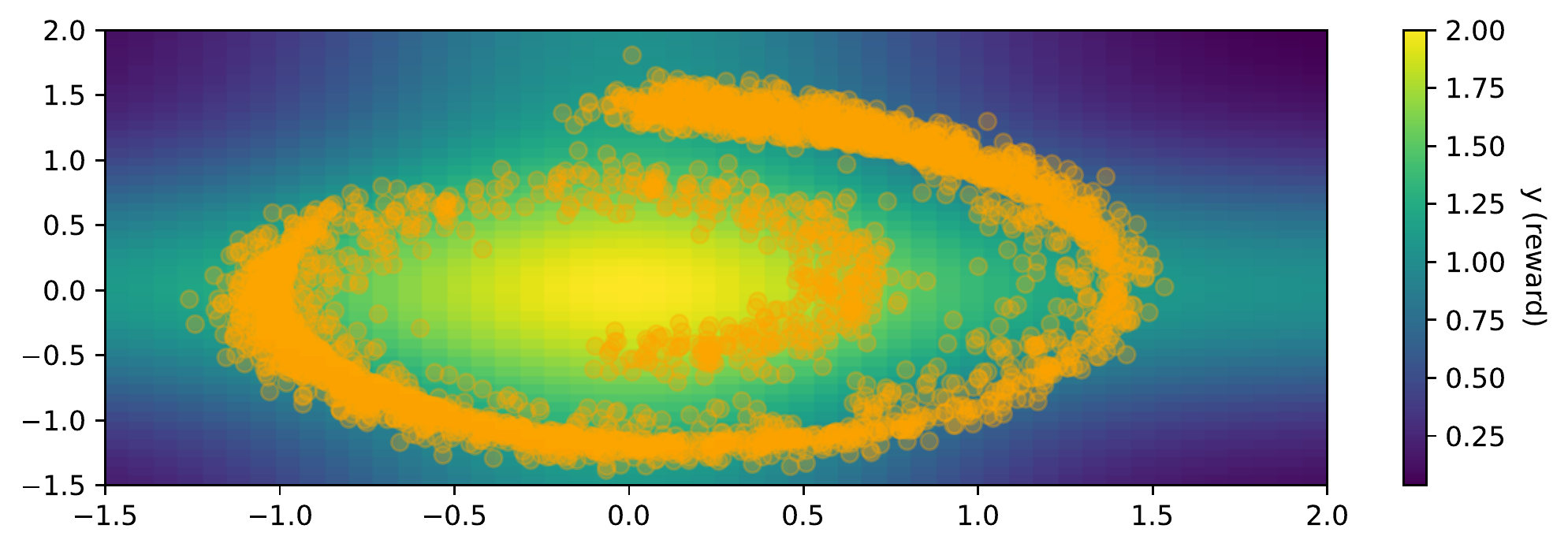}
    \caption{2D spiral dataset. Orange points are samples from the ground truth marginal $p(\xx)$, and background colours correspond to values of $y$ for the ground truth oracle $f(\xx) = \sum_{i=1}^2 \exp( (\xx_i - 0)^{2} )$.}
    \label{fig:dataset}
\end{figure}

\paragraph{Dataset} We consider a simple 2D spiral dataset that has been modified to also introduce a reward variable $y$. The ground truth function for this reward variable is $f(\xx) = \sum_{i=1}^2 \exp( (\xx_i - 0)^{2} )$, which means that the largest reward is found at the origin $(\xx_1, \xx_2) = (0,0)$. This is illustrated in Figure \ref{fig:dataset}. In the context of MBO, we would like to learn a generative model which is able to sample \emph{valid} points that are as close to the center as possible, since points that are closest to the center will have a larger reward. Here, a `valid' point is one that lies on the spiral, i.e. some $\xx$ for which $p(\xx) > 0$ for the ground truth distribution $p(\xx)$. As we can see in the figure, the point $\xx = (\xx_1, \xx_2) = (0,0)$ which lies at the center would not be valid.

\paragraph{Training} The energy model is a one hidden layer MLP of 256 units, and we use $k$-contrastive divergence during training for $k=100$ time steps (see Equation \ref{eq:cd_k}). Its associated variance schedule for Langevin MCMC is a geometric sequence from $0.02 \rightarrow 0.001$ over those $k$ intervals.\footnote{This can be generated easily in Numpy via \texttt{numpy.geomspace(a=0.02, b=0.001, n)}}. At generation time, we run Langevin MCMC for 50k timesteps with prior distribution $p_{\pi}(\xx_0) = \text{Uniform}(-1.5, 2)$ and use a geometric schedule from $0.1 \rightarrow 10^{-5}$.

\begin{figure}[h!]
    \centering 
    \begin{subfigure}[b]{0.45\textwidth}
        \centering
        \includegraphics[page=1,width=\textwidth]{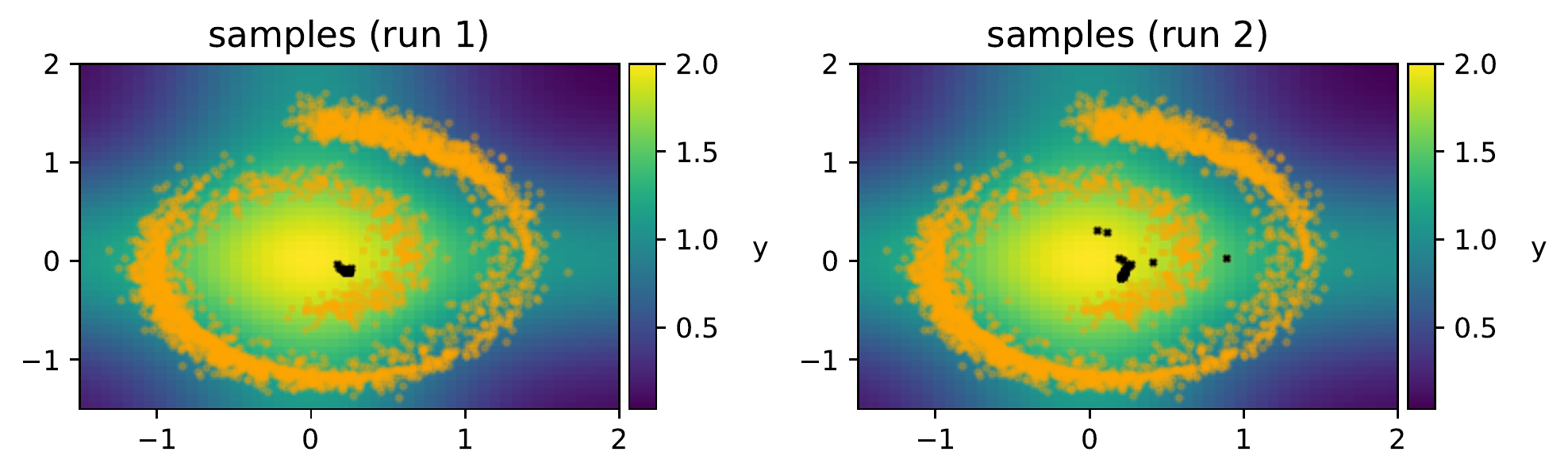}
        \caption{\textit{\textbf{Original COMs}}: $\alpha = 0$}
        \label{fig:samples_orig_0}
    \end{subfigure}
    \begin{subfigure}[b]{0.45\textwidth}
        \centering
        \includegraphics[page=1,width=\textwidth]{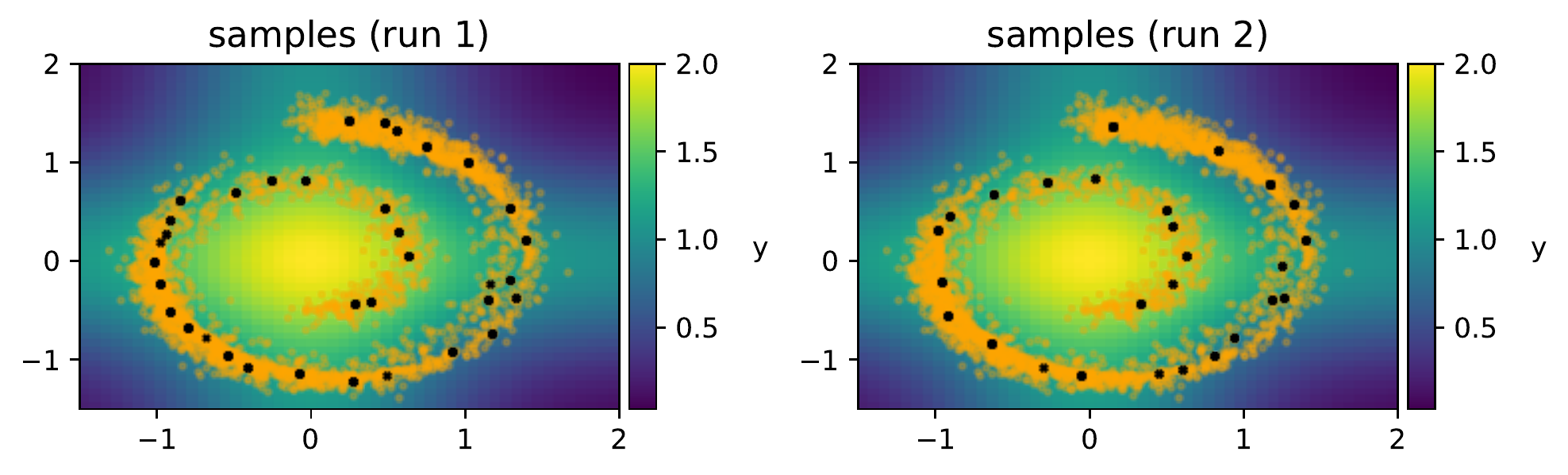}
        \caption{$\alpha = 50$}
        \label{fig:samples_orig_50}
    \end{subfigure}
    \caption{Generated samples (shown as black crosses) for the \textit{\textbf{original COMs}} formulation (Sec. \ref{sec:coms}). Orange points are those from the real distribution $p(\xx)$, and the colourbar denotes the ground truth $y$. In \ref{fig:samples_orig_0} and \ref{fig:samples_orig_50} we show samples for $\alpha = 0$ and $\alpha = 50$ respectively, for two separate training runs (seeds). For $\alpha = 0$, only the conditional distribution $\pt(y|\xx)$ modelled, rather than $\xx$ and $y$ jointly. For the $\alpha = 50$ case, the energy loss is heavily weighted in favour of modelling $\pt(\xx)$. Because the original COMs formulation uses a gradient ascent MCMC sampler, only modes can be sampled from the distribution, and sample diversity suffers as a consequence. This issue is addressed with \textit{\textbf{Stochastic COMs}} (Fig. \ref{fig:samples_stochastic}), which uses Equation \ref{eq:langevin_mcmc_coms_uncond} to properly sample from the distribution.}
    \label{fig:samples_orig}
\end{figure}

\begin{figure}[h!]
    \centering 
    \begin{subfigure}[b]{0.45\textwidth}
        \centering
        \includegraphics[page=1,width=\textwidth]{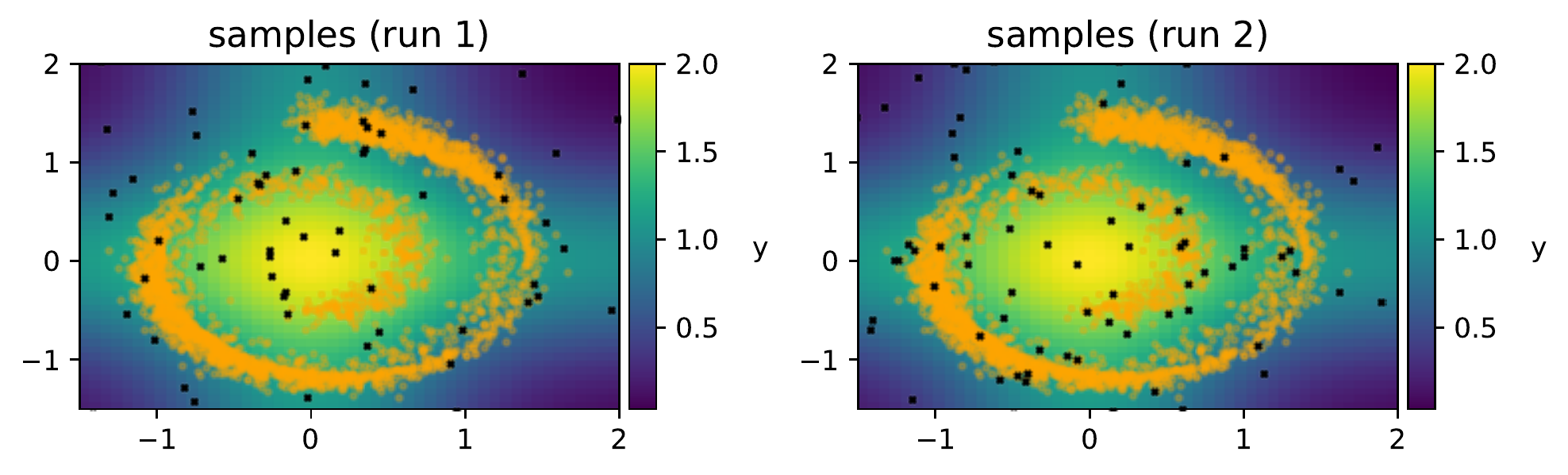}
        \caption{\textit{\textbf{Stochastic COMs}}: $\alpha = 0$}
        \label{fig:samples_stochastic_0}
    \end{subfigure}
    \begin{subfigure}[b]{0.45\textwidth}
        \centering
        \includegraphics[page=1,width=\textwidth]{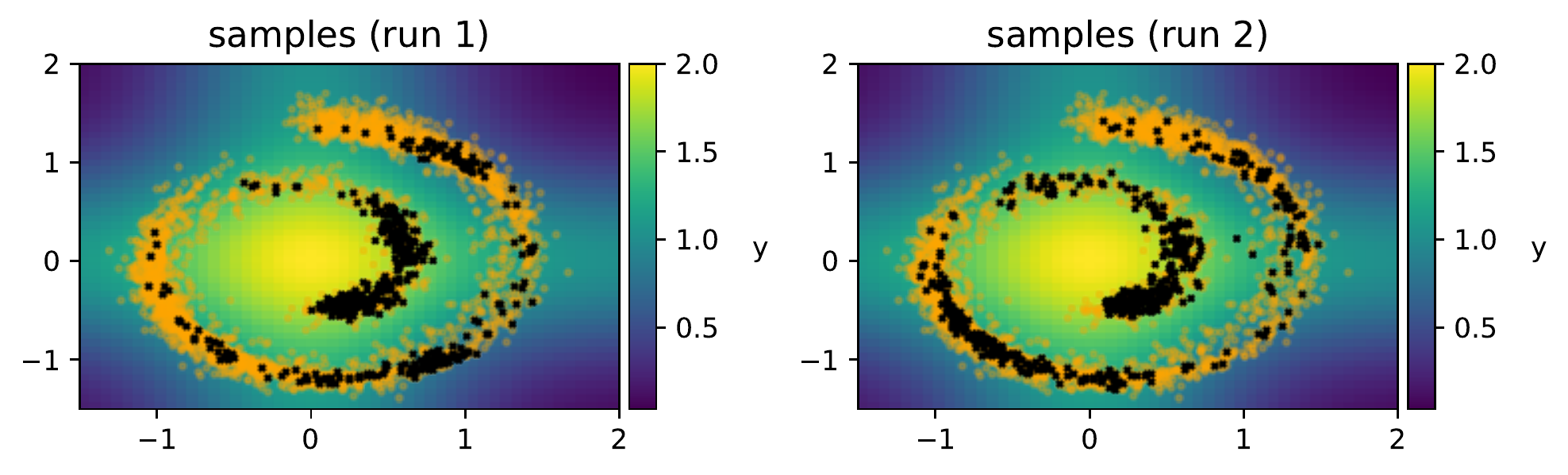}
        \caption{$\alpha = 50$}
        \label{fig:samples_stochastic_50}
    \end{subfigure}
    \caption{Generated samples (shown as black crosses) for the \textit{\textbf{stochastic COMs}} formulation (Sec. \ref{sec:coms_stochastic}). Orange points are those from the real distribution $p(\xx)$, and the colourbar denotes the ground truth $y$. In \ref{fig:samples_stochastic_0} and \ref{fig:samples_stochastic_50} we show samples for $\alpha = 0$ and $\alpha = 50$ respectively, for two separate training runs (seeds). For $\alpha = 0$, only the conditional distribution $\pt(y|\xx)$ modelled, rather than $\xx$ and $y$ jointly. For the $\alpha = 50$ case, the energy loss is heavily weighted in favour of modelling $\pt(\xx)$. While samples for both $\alpha$'s are more diverse, it is difficult to select for the `good' samples, i.e. those that are close to the center while still lying on the spiral (see Figure \ref{fig:com_stochastic_additional}) for additional enumerations of $\alpha$). This is because the same energy function is being used to parameterise both distributions. We resolve this issue with the decoupled COMs variant, which is shown in Figure \ref{fig:samples_dc}.}
    \label{fig:samples_stochastic}
\end{figure}

\begin{figure}[h!]
    \centering 
    \begin{subfigure}[b]{0.32\textwidth}
        \centering
        \includegraphics[page=1,width=\textwidth]{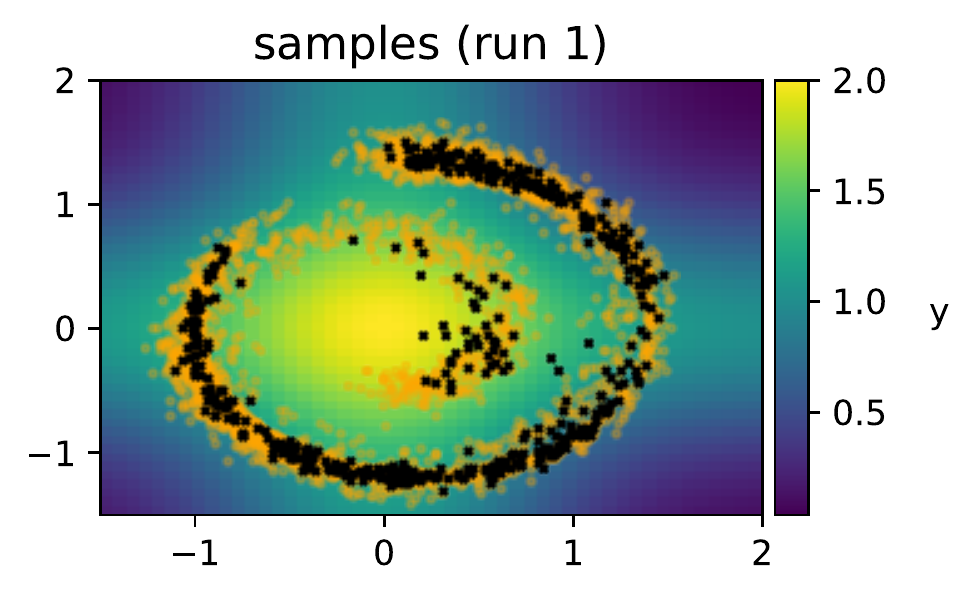}
        \caption{\textit{\textbf{Decoupled COMs}}: $w = 0$}
        \label{fig:samples_dc1}
    \end{subfigure}
    \begin{subfigure}[b]{0.32\textwidth}
        \centering
        \includegraphics[page=1,width=\textwidth]{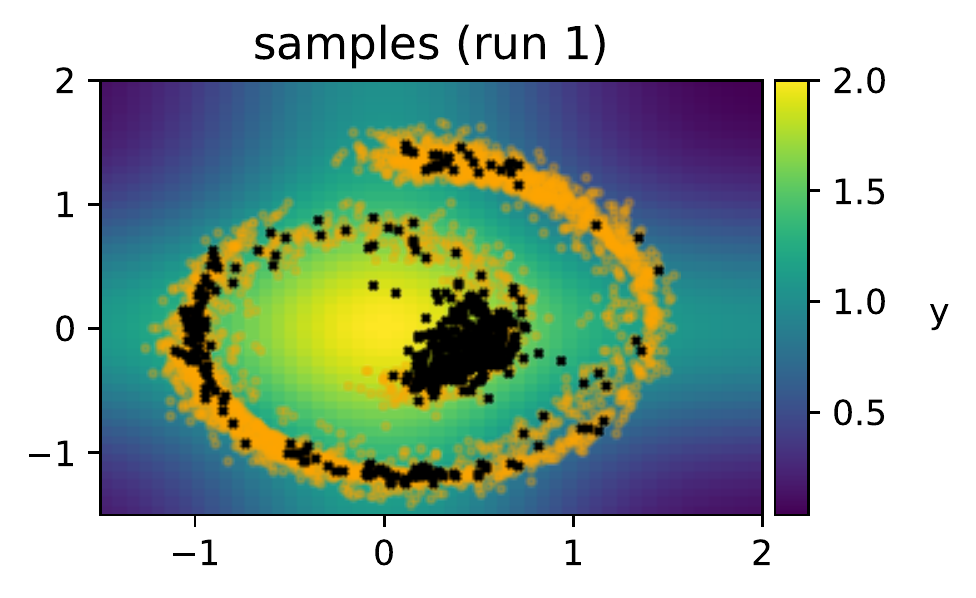}
        \caption{$w = 5$}
        \label{fig:samples_dc2}
    \end{subfigure}
    \begin{subfigure}[b]{0.32\textwidth}
        \centering
        \includegraphics[page=1,width=\textwidth]{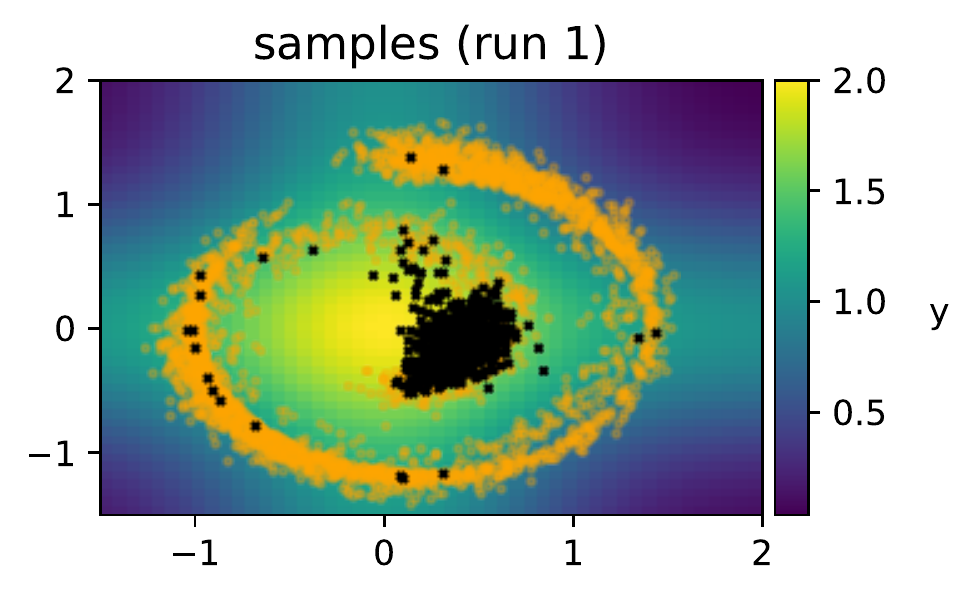}
        \caption{$w = 10$}
        \label{fig:samples_dc3}
    \end{subfigure}
    \caption{Generated samples (shown as black crosses) for \emph{\textbf{decoupled COMs}} (Sec. \ref{sec:decoupled}). Like stochastic COMs, Langevin MCMC is also used here but we also leverage the gradient of an externally trained regression model $\fomega(\xx)$ as shown in Equation \ref{eq:langevin_mcmc_decoupled}. Here, we achieve the desired behaviour: a modest value of $w$ gives us samples that mostly lie on the part of the spiral closest to the center.}
    \label{fig:samples_dc}
\end{figure}

\paragraph{Results} We train each of the three variants of COMs, and their results are shown in Figures \ref{fig:samples_orig}, \ref{fig:samples_stochastic}, and \ref{fig:samples_dc}, respectively. For the first two, we train two variants: one where $\alpha = 0$ and the model reduces down to just a regression model (classifier), and one where $\alpha = 50$ where the model is heavily weighted to model the data distribution instead. Since the original COM uses gradient ascent as its sampler, samples are heavily biased towards seeking modes and sample diversity suffers as a result. In the stochastic variant this is fixed, however we found it difficult to choose an $\alpha$ such that samples were simultaneously concentrated near the center but \emph{on} the spiral, which would constitute the best samples for this dataset. As we mentioned in Section \ref{sec:connection}, we believe it is because we're using the same energy model to model both $\pt(\xx)$ and $\ft(\xx)$, and therefore either task is not able to be learned sufficiently well. In decoupled COMs however (Figure \ref{fig:samples_dc}) the energy $\Et(\xx)$ and $\fomega(\xx)$ are separate models and the latter is weighted by hyperparameter $w$. We can see that for modest values of $w$ we obtain samples that progressively become more heavily concentrated at the center, but are still lying on the spiral.


      



\newcommand{\minipagewidth}{0.4}
\begin{table*}[]
    \footnotesize
    \centering
    \caption{Summary of the three variants of COMs in this work. \emph{Stochastic} (Sec. \ref{sec:coms_stochastic}) and \emph{decoupled} (Sec. \ref{sec:decoupled}) variants are proposed methods that address particular issues in the original formulation. Note that for all generated samples shown in Figures \ref{fig:samples_orig}, \ref{fig:samples_stochastic} and \ref{fig:samples_dc}, either $\pt(\xx)$ is used (original, \emph{stochastic}) or the exponentially tilted variant $\pt(\xx)\exp(\fomega(\xx)^{w})$ (\emph{decoupled}). $\dagger$ = samples converge to a maximum a posteriori solution, so we are not truly sampling from $\pt(\xx)$ (this is addressed in the stochastic variant).}
    \label{tb:metrics}
    \vspace{0.1cm}
    \begin{tabular}{p{2.5cm}p{5.7cm}p{4cm}p{2.5cm}}
    \toprule
    Method                     & Joint density & Training algorithm for $\pt(\xx)$  & Sampling algorithm for $\pt(\xx)$  \\
    \midrule
    COMs (Sec. \ref{sec:coms}) & 
    \begin{minipage}{\minipagewidth\textwidth}\vspace{0pt}
        $\pt(\xx, y; \alpha) \propto \pt(\xx)^{\alpha}\pt(y|\xx)$
    \vspace{0pt}\end{minipage} & Contrastive divergence (approximate$^\dagger$) (Eqn \ref{eq:coms_loss}) & Gradient ascent$^{\dagger}$ (Eqn. \ref{eq:grad_asc}) \\
    \midrule
    Stochastic COMs (Sec. \ref{sec:coms_stochastic}) & 
    \begin{minipage}{\minipagewidth\textwidth}\vspace{0pt}
        $\pt(\xx, y; \alpha) \propto \pt(\xx)^{\alpha}\pt(y|\xx)$
    \vspace{0pt}\end{minipage} & Contrastive divergence & Langevin MCMC (Eqn. \ref{eq:langevin_mcmc_coms_uncond}) \\
    \midrule
    Decoupled COMs (Sec. \ref{sec:decoupled}) & 
    \begin{minipage}{\minipagewidth\textwidth}\vspace{0pt}
        $p_{\theta,\omega}(\xx, y; w) \propto \pt(\xx)\exp(\fomega(\xx)^{w})\pt(y|\xx)$
    \vspace{0pt}\end{minipage} & Contrastive divergence & Langevin MCMC (Eqn. \ref{eq:langevin_mcmc_decoupled})
    \end{tabular}
\end{table*}

\section{Related work}

Recently, \emph{score-based generative models (SBGMs)} have been in wide use \cite{song2019generative, song2020improved}, and this also includes the diffusion class of models since they are theoretically very similar \cite{sohl2015deep, ho2020denoising}. Due to space constraints we defer an extended discussion to Section \ref{sec:sms}, though we heavily conjecture that this class is model is significantly more robust than COMs and therefore CD-EBMs. This is for the following reasons:
\begin{itemize}
\item SBGMs sidesteps the issue of having to generate samples from the distribution during training with MCMC, which significantly speeds up training. This is because the training objective used is score matching (matching derivatives), as opposed to contrastive divergence which requires negative samples be generated.
\item SBGMs model the gradient directly $\st(\xx) = \nabla_{\xx} \log \pt(\xx)$. Not only does this bypass the need to compute gradients at generation time with autograd, it also means that the energy function can model more information about its input because it is now a mapping from $\mathbb{R}^{d} \rightarrow \mathbb{R}^{d}$ (where $d$ is the input data dimension), as opposed to $\Et(\xx)$ which is a mapping from $\mathbb{R}^{d} \rightarrow \mathbb{R}$ \citep{salimans2021should}. Furthermore, this mapping from $\mathbb{R}^{d} \rightarrow \mathbb{R}^{d}$ allows one to use specialised encoder-decoder models such as the U-Net \citep{ronneberger2015u}, which leverages skip connections to combine information at various resolutions of the input. 
\item Modern SBGMs also propose score matching over \emph{many} different noise scales. Both large and small are important, since larger ones make it easier to cover all modes and smaller ones are closer to the score of the actual data distribution. All of these noise scales are learned within the same network $\st(\xx)$. At generation time, these noise scales are combined to give rise to an annealed version of Langevin MCMC which iterates from larger noise scales to smaller ones.
\end{itemize}

We note that a modern SBGM can be constructed by simply replacing the contrastive-based formulation of $\pt(\xx)$ in the decoupled COMs variant (Section \ref{sec:decoupled}) with one that has been trained with score matching as per \citet{song2019generative}. This combined with an external classifier $\fomega$ becomes very reminiscent to the `classifier guidance' style of techniques introduced in \citet{dhariwal2021diffusion, ho2022classifier}.

\section{Conclusion}

In this work, we showed that COMs, a highly performant algorithm for offline model-based optimisation, is essentially an energy-based model trained via contrastive divergence. COMs use the same energy model to parameterise both the unconditional and conditional parts of the data distribution ($\pt(\xx)$ and $p(y|\xx)$, respectively), and this also means that the model has a special property in which the probability of sampling an input is \emph{proportional} to its predicted reward. In this work we identified two shortcomings with the original formulation: firstly, a gradient ascent sampler is used which limits sample diversity; and secondly the parameterisation of both distributions hinders conditional sampling quality, as demonstrated on a toy 2D spiral dataset. We address both of these issues with a `decoupled' variant of COMs which models the conditional and unconditional parts of the joint distribution separately, as well as use a Langevin MCMC sampler which correctly samples from the learned distibution. Lastly, we contribute a brief discussion comparing the training dynamics of COMs with more recent energy-based models which are trained with score matching.

\bibliography{paper}
\bibliographystyle{icml2023}


\makeatletter
\renewcommand \thesection{S\@arabic\c@section}
\renewcommand\thetable{S\@arabic\c@table}
\renewcommand \thefigure{S\@arabic\c@figure}
\makeatother

\newpage
\appendix
\onecolumn
\section{Appendix}

\subsection{Additional figures}

\begin{figure}[h]
    \centering 
    \begin{subfigure}[b]{0.45 \textwidth}
        \centering
        \includegraphics[page=1,width=\textwidth]{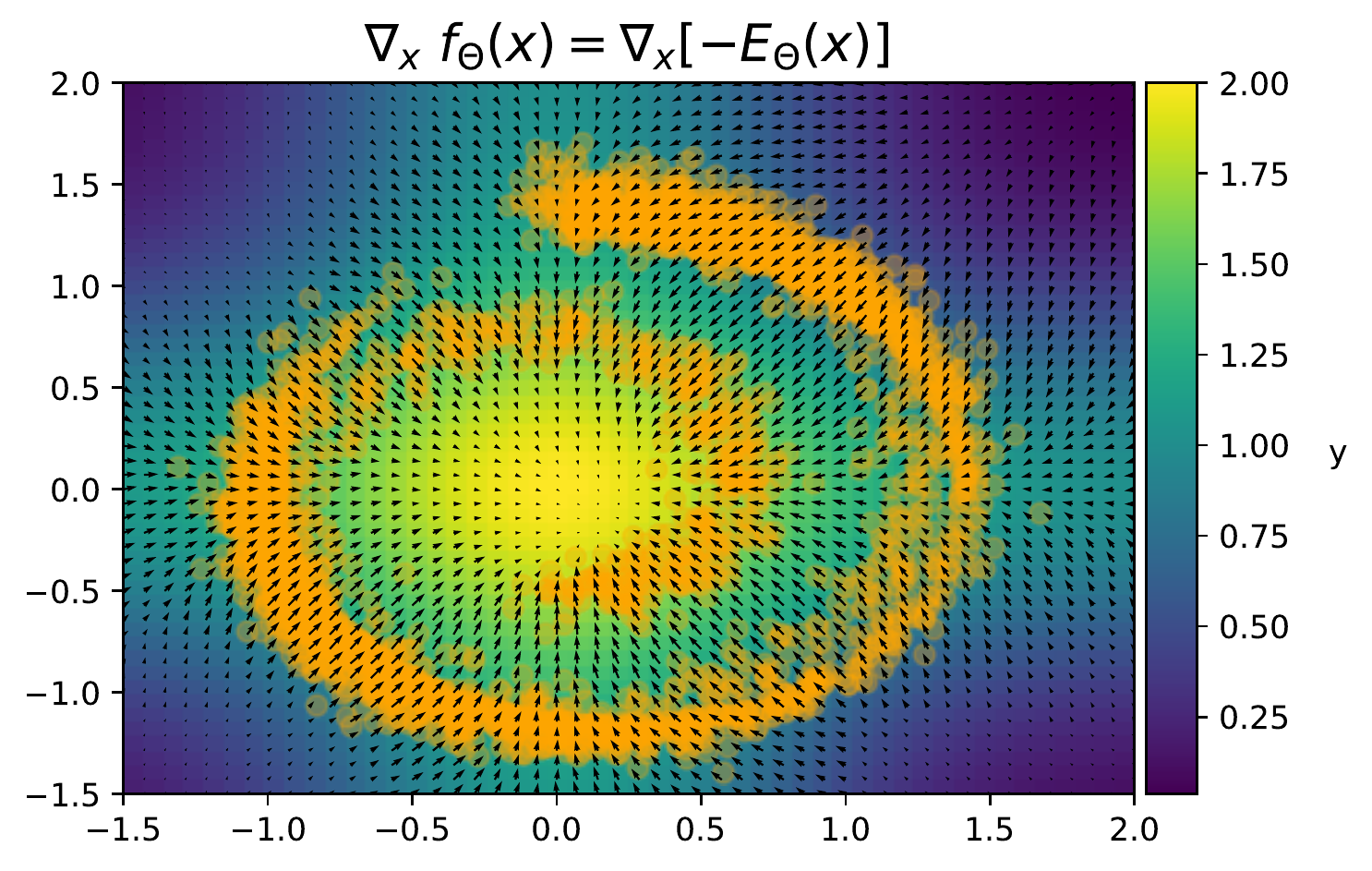}
        \caption{$\alpha = 0$}
        \label{fig:alpha0}
    \end{subfigure}
    \begin{subfigure}[b]{0.45\textwidth}
        \centering
        \includegraphics[page=1,width=\textwidth]{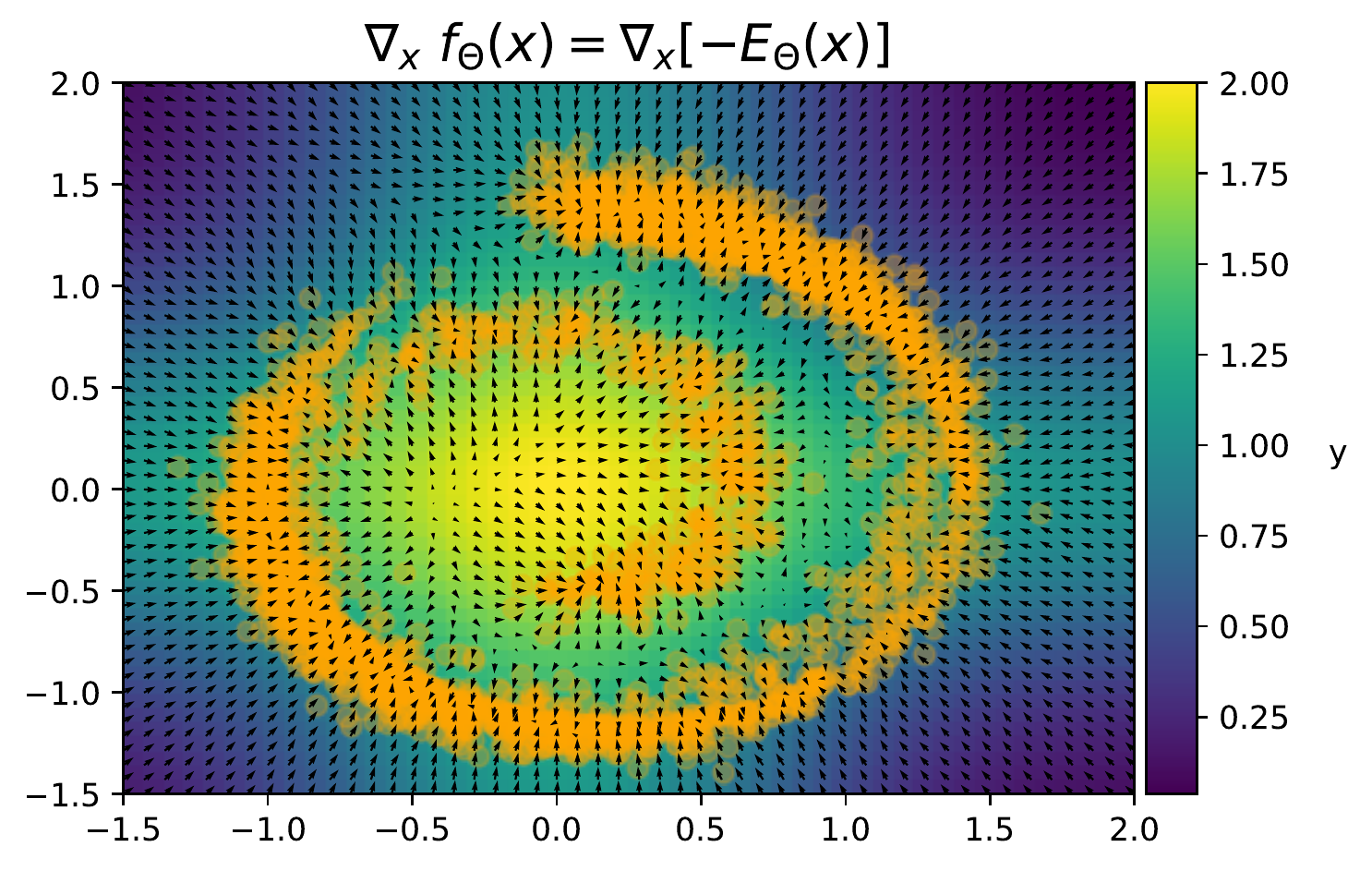}
        \caption{$\alpha = 100$}
        \label{fig:alpha50}
    \end{subfigure}
    \caption{Stochastic COMs (Section \ref{sec:coms_stochastic}). Vector field plots for the learned energy function, for $\alpha = 0$ (\ref{fig:alpha0}) and $\alpha = 50$ (\ref{fig:alpha50}). Best viewed with a PDF viewer at a higher zoom level. Training details are specified in Section \ref{sec:toy}.}
    \label{fig:quiver_plots}
\end{figure}

We can visualise the gradient of the energy function $\nabla_{\xx}[-\Et(\xx)] = \nabla_{\xx}\ft(\xx)$ by plotting their values (vectors) over the entire 2D grid, creating a vector field. This is shown in Figure \ref{fig:quiver_plots}. We can see that when $\alpha = 0$, the energy function points to the center of the spiral since it is only trained to predict the reward of $\xx$ and this is where the predicted reward is largest. Conversely, when $\alpha = 50$ the energy model is heavily weighted in favour of modelling the (unconditional) distribution of $\xx$ during training, and the vector field points towards examples \emph{on} the spiral. 

\begin{figure}[h]
    \centering
    \includegraphics[page=1,width=0.8\textwidth]{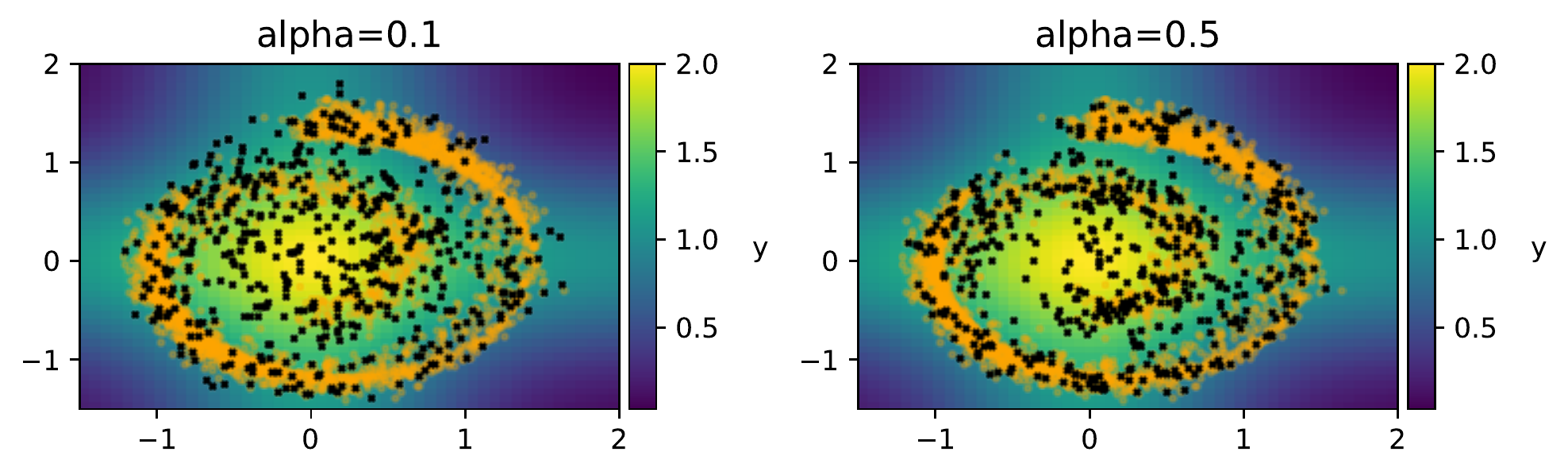}
    \includegraphics[page=1,width=0.8\textwidth]{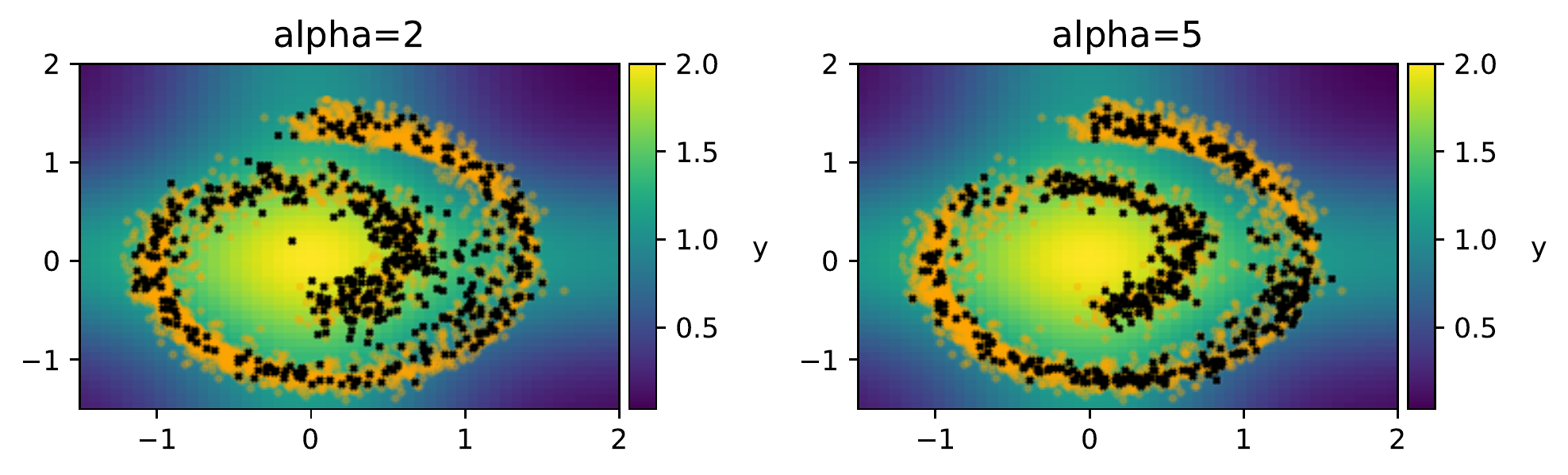}
    \caption{Additional plots to complement Figure \ref{fig:samples_stochastic}. Values of $\alpha$ are denoted in each subplot. Similar to Figure \ref{fig:samples_stochastic}, even with small values of $\alpha$ we are unable to generate samples such that they both lie on the spiral \emph{and} are close to the center, where the ground truth reward $y$ is largest.}
    \label{fig:com_stochastic_additional}
\end{figure}

\subsection{Additional math}

This proof is used for Equation \ref{eq:integral_trick}. From Thereom 6 of \citet{imai2006pol} we have that, satisfying some conditions, the derivative of the expected value is the expected value of the derivative:
\begin{align}
\nabla_{\theta} \big[ \mathbb{E}_{\xx \sim \pt(\xx)} \Et(\xx) \big] & = \nabla_{\theta} \big[ \int_{\xx} \pt(\xx) \Et(\xx) d\xx \big] \\
& = \int_{x} -\nabla_{\xx} \Et(\xx) \pt(\xx) d\xx \ \ \text{(thm. 6)} \\
& = \mathbb{E}_{\xx \sim \pt(\xx)} \big[ -\nabla_{\xx} \Et(\xx) \big].
\end{align}

\subsection{Score-matching EBMs (SM-EBMs)} \label{sec:sms}

A recent class of generative model that has enjoyed immense success is the \emph{score-matching EBM} (SM-EBM). Score matching refers to the minimisation of the \emph{Fisher} divergence between the real and generative distributions, which is equivalent to measuring the difference between their respective log derivatives \cite{hyvarinen2005estimation}:
\begin{align} \label{eq:fisher}
D_{F}( p(\xx) || \pt(\xx) ) = \mathbb{E}_{\xx \sim p(\xx)}\Big[ \frac{1}{2} \| \nabla_{\xx} \log p(\xx) - \nabla_{\xx} \log \pt(\xx) \|^{2} \Big].
\end{align}
Here, the score refers to the the derivative of the log density with respect to the input, i.e. $\nabla_{\xx} \log \pt(\xx)$. As we have seen already, this can be parameterised with an energy model $\Et(\xx)$. Unfortunately, in its current form Equation \ref{eq:fisher} is intractable since it is assumed the score of the data distribution is known. While an equivalent form can be written that only relies $\pt(\xx)$, it relies on computing the Hessian and has quadratic time complexity in the dimension of $\xx$ \cite{hyvarinen2005estimation}. To address these issues (and other theoretical assumptions about the $p(\xx)$), denoising score matching \cite{vincent2011connection} was proposed, where the Fisher divergence is computed with respect to a `noisy' version of the data distribution, $q(\xx)$. This can be expressed as the following marginalisation over a conditional noising distribution $q_{\sigma}(\xx|\xx_0)$ and the actual data distribution $\pt(\xx)$:
\begin{align} 
q_{\sigma}(\xx) = \int_{\xx_0} q_{\sigma}(\xx|\xx_0)p(\xx_0) d\xx, 
\end{align}
where $q_{\sigma}(\xx|\xx_0) = \mathcal{N}(\xx; \xx_0, \sigma^2)$. While this no longer becomes an estimate of $D_F$ when $p(\xx)$ is replaced with $q(\xx)$, the difference is negligible for small values of $\sigma$. 

Recently, \emph{score-based generative models (SBGMs)} were proposed \cite{song2019generative, song2020improved}, which can be thought of as an improved version of the denoising score matching EBM but with the added intention of generating samples, which comes in the form of a modified Langevin MCMC sampler. To avoid a deluge of acronyms and terms, let us simply refer to these as `modern' SM-EBMs. Modern SM-EBMs are currently state-of-the-art, in no small part due to some tricks which improve the training dynamics of the original score matching algorithm \cite{vincent2011connection}. Firstly, instead of modelling an energy function and then having to backprop through it to obtain the actual score, the score is parameterised directly, i.e. $\st(\xx) \approx \nabla_{\xx} \log p(\xx)$ instead of $\nabla_{\xx}[-\Et(\xx)] \approx \nabla_{\xx} \log p(\xx)$. This means that $\st : \mathbb{R}^{d} \rightarrow \mathbb{R}^{d}$ and encoder-decoder-style architectures must be used. Secondly, denoising score matching has issues with modelling modes of the data distribution when they are separated by low (or zero) density regions. While larger noise perturbations make this easier, the noise distribution $q(\xx)$ would become less representative of the actual data distribution $p(\xx)$. To resolve this dilemma, a series of noise distributions are used instead. For some $t \in \{1, \dots, T\}$:
\begin{align}
q_t(\xx|\xx_0) & = q_{\sigma_t}(\xx|\xx_0) = \mathcal{N}(\xx; \xx_0, \sigma_t^2) \\
\implies q_t(\xx) & = \int_{\xx_0} q_{\sigma_t}(\xx|\xx_0) p(\xx_0) d\xx_0,
\end{align}
where the sequence of $\sigma_t$'s follows a positive geometric sequence, and $T$ is large enough such that $\sigma_T \approx 0$ (so $q_T \approx p(\xx)$). With this in mind, we must also modify the score predictor to also condition on a timestep, i.e. $\st(\xx; t)$. $\st(\xx; t)$ is trained to estimate the score $\nabla_{\xx} \log q_t(\xx)$. We defer training details to \citet{song2019generative, song2020improved}, though it suffices to say that at generation time an annealed version of Langevin MCMC is used where the noise magnitude $\sigma$ progressively becomes smaller as the number of timesteps increases.

\end{document}